# An enhanced motion planning approach by integrating driving heterogeneity and long-term trajectory prediction for automated driving systems

Ni Dong[1], Shuming Chen[1], Yina Wu[2], Yiheng Feng[3], Xiaobo Liu[1*]

[1] School of Transportation and Logistics, Southwest Jiaotong University, Chengdu 611756, China

[2] Department of Civil, Environmental & Construction Engineering, University of Central Florida, FL, United States

3 Lyles School of Civil Engineering, Purdue University, West Lafayette, IN, United States

Corresponding author: xiaoboliu@swjtu.cn

**Abstract** Navigating automated driving systems (ADSs) through complex driving environments is difficult. Predicting the driving behavior of surrounding human-driven vehicles (HDVs) is a critical component of an ADS. This paper proposes an enhanced motion-planning approach for an ADS in a highway-merging scenario. The proposed enhanced approach utilizes the results of two aspects: the driving behavior and long-term trajectory of surrounding HDVs, which are coupled using a hierarchical model that is used for the motion planning of an ADS to improve driving safety. An unsupervised clustering algorithm is utilized to classify HDV drivers into two categories: aggressive and normal, as part of predicting their driving behaviors. Subsequently, a logistic regression model is employed for driving style prediction. For trajectory prediction, a transformer-based model that concentrated parallelization computations on longer sequence predictions via a self-attention mechanism is developed. Based on the predicted driving styles and trajectories of the surrounding HDVs, an intelligent decision-making strategy is utilized for ADS motion planning. Finally, real-world traffic data collected from drones on a highway ramp in Xi'an is used as a case study to train and evaluate the proposed model. The results demonstrate that the proposed approach can predict HDVs merging trajectories with a mean squared error (MSE) smaller than 0.05 at thirty seconds away from the merging point, outperforming existing approaches in terms of predictable duration and accuracy. Furthermore, it exhibited safety improvements by adjusting the ADS motion state in advance with good predictive power for the surrounding HDVs' motion.

# 1. Introduction

Automated driving systems (ADSs) have significant potential for improving public safety on roadways (Claybrook and Kildare, 2018). The benefits of ADSs can be guaranteed by making the driving experience in complex driving environments more comfortable and safer (Sarker et al., 2019). New perception technologies enable ADSs to detect the surrounding traffic. When surrounding traffic, such as other vehicles, pedestrians, and cyclists, is detected, a motion-planning algorithm can generate a safe path for the ADS (Frazzoli, 2000; Shiller and Gwo, 1991). The generated path is continuously updated using decision and control technologies based on the surrounding environment.

One of the greatest challenges for ADSs is the uncertainty of the surrounding dynamic environment (González et al., 2016). An ADS must adapt to these changing conditions and make decisions that prioritize safety and efficiency. However, relying solely on perceiving the surrounding dynamic changes has some limitations, including perceptual uncertainty owing to the complexity of scenarios and dynamic occlusion between vehicles (Czarnecki and Salay, 2018; Van Brummelen et al., 2018; Chen et al., 2019; Gilroy et al., 2019; Cao et al., 2022; Saad et al., 2022), and time consumption owing to the heavy computational burden in implementing object detection and tracking algorithms (Gandhi and Trivedi, 2007; Su et al., 2023). Addressing these limitations is the key to the ability of an ADS to react adequately to dynamic environments, particularly in urban scenarios where multiple agents (other vehicles, pedestrians, and cyclists) must be considered. The uncertainty of the perception of the ADS can be overcome by predicting the future motion of the surrounding traffic. By anticipating the future states of other vehicles, pedestrians, and objects in the environment, an ADS can plan its movements accordingly and avoid potential collisions to ensure safe operation (Frazzoli, 2000; Shiller and Gwo, 1991). However, inaccurate prediction may lead to an incorrect judgement of the future states of traffic participants. Another challenge in ADS motion planning is the time limitation for generating a new collision-free trajectory with multiple dynamic obstacles (González et al., 2016). This is primarily because ADS perception is time-consuming, which reduces the available time for motion planning (Gandhi and Trivedi, 2007). Therefore, a timely ADS motion planning with the predicted surrounding traffic states is critical to operate with up-to-date information about the environment.

Recent advances in traffic state prediction have captured the interest of academics and industries. In this paper, we focus on predicting the motion of surrounding human-driven vehicle (HDV) of an ADS. The prediction of HDV motion can be broadly categorized into two types: behavior and trajectory prediction. For human driving behavior prediction, significant interest and effort have been invested in detecting and predicting the heterogeneity of driving behavior. This refers to the observation that different individuals may exhibit varying driving styles and habits even under similar driving scenarios (Ding et al., 2022;

Makridis et al., 2020). Recent research has focused on developing methods for identifying and understanding the intent of human drivers (Makridis et al., 2020; Ding et al., 2022; Carrone et al., 2021), enabling ADSs to adapt to their driving styles (Schwarting et al., 2019). For example, an ADS can be programmed to decelerate in advance when approaching a driver who tends to brake suddenly. The other critical aspect of HDV motion prediction, trajectory prediction, has attracted considerable interest in recent years. The accuracy of trajectory prediction has been significantly enhanced by leveraging the advancements in machine learning technology (Doshi and Trivedi, 2011). Deep learning models, such as recurrent neural networks (RNNs) and their derivatives, have proven to achieve higher accuracy than traditional machine learning models for trajectory prediction. This is because they can capture complex patterns and relationships within data (Dang et al., 2017; Xing et al., 2020).

These prediction technologies can anticipate the future movements of surrounding HDVs. The generated path is updated continuously using decision strategies that consider the trajectory predictions of the surrounding HDVs (González et al., 2016). This approach can improve the safety of ADSs by using pre-generated trajectories. Most studies have focused on short-term vehicle trajectories using RNN and long short-term memory (LSTM) methods (Lee and Kum, 2019). However, these methods have certain limitations, including being computationally expensive to train and memory limitations for long sequences (Giuliari et al., 2021). Compared with short-term vehicle trajectory prediction, long-term vehicle trajectory prediction can help autonomous vehicles anticipate the actions of other vehicles over a longer period, thereby reducing the risk of accidents (Shiller and Gwo, 1991). However, few studies have focused on long-term trajectory predictions. This study aimed to fill this gap in the literature.

In terms of ADS motion planning, the existing literature primarily focuses on making decisions and instructing an ADS to react accordingly to avoid incidents (Claussmann et al., 2019; González et al., 2015; Katrakazas et al., 2015). These motion-planning approaches include four aspects: graph search, sampling, interpolation, and numerical optimization (Gonzalez et al., 2015). Various motion-planning technologies have been applied to complex environments. However, most studies focused on generating a safe trajectory by avoiding static obstacles (Bounini et al., 2017; Ji et al., 2017). These motion-planning algorithms are often inadequate for handling dynamic changes in the surrounding environment. However, insufficient research has been conducted on ADS motion planning that considers both human driving heterogeneity and the long-term trajectory prediction of surrounding vehicles. The link between the motion prediction of surrounding vehicles and ADS motion planning can enable the planning process to prevent dangerous scenarios (Huang et al., 2020).

To address these problems, we propose an enhanced ADS motion-planning framework for highway merging scenarios. The main contributions of this paper are as follows. First, a hierarchical model to predict the driving heterogeneity and trajectory of surrounding HDVs is proposed. Specifically, an unsupervised clustering algorithm is used to group merging HDV driving heterogeneity, whereas different surrounding HDV driving styles are predicted using a logistic regression method. Thus, a transformer-based trajectory-prediction model is developed. Second, the hierarchical model has effective prediction, enabling the ADS to obtain the long-term trajectories of surrounding HDVs in advance. Our trajectory-prediction method is superior to existing methods in terms of prediction accuracy and horizon. Third, an enhanced motion-planning framework is proposed by integrating the hierarchical model, ensuring the safety improvement of the ADS by dynamically adjusting the decision-making strategies (i.e., cruise, following, real-time avoidance, and avoidance in advance). The optimal prediction horizon is identified using the time-to-collision (TTC) between an ADS and HDV. Fig. 1 shows the research framework.

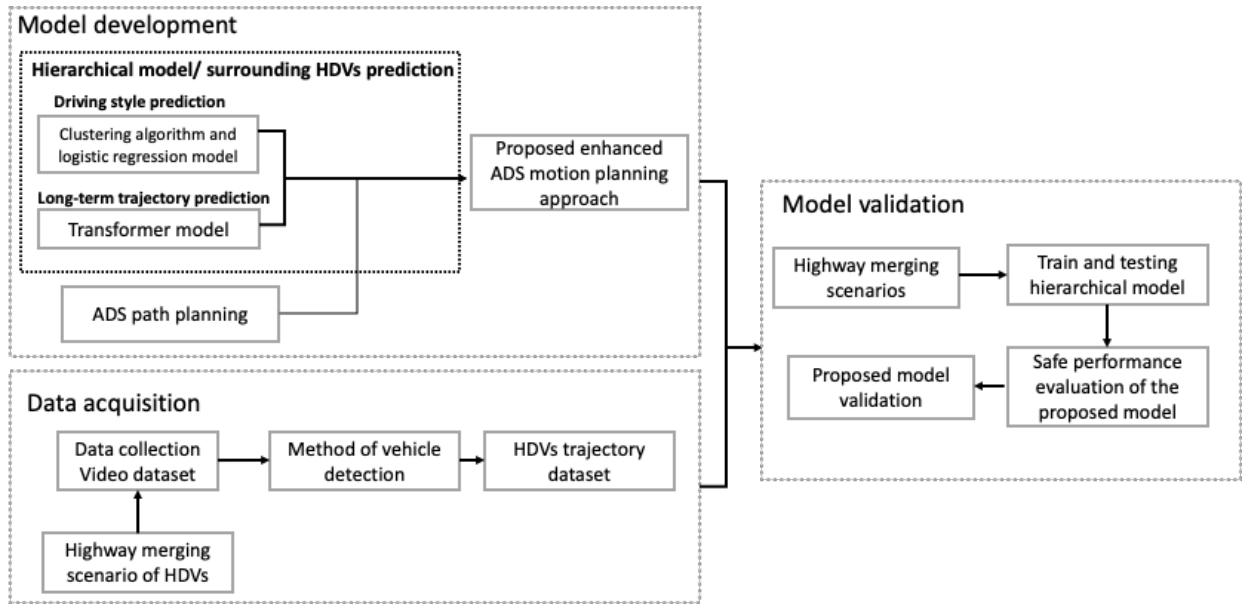

**Fig. 1.** Research framework

The remainder of this paper is organized as follows. Section 2 provides a literature review on two major approaches of surrounding HDVs hierarchical model (i.e., driving style and trajectory prediction) and ADS motion planning. Section 3 describes the research scenario. Section 4 introduces the proposed enhanced motion-planning approach, including HDV hierarchical and ADS motion-planning models. Section 5 describes a case study to demonstrate the performance of the proposed approach. Section 6 concludes the paper and provides potential research directions.

## 2. Related studies
### 2.1 Detection and prediction of HDV driving heterogeneity

Multiple human drivers can cause heterogeneity in populations and systems. Heterogeneity can be caused by differences in the physical, genetic, cultural or social characteristics of individuals within a population. Several methods are available for detecting and predicting driving heterogeneity; among them, rule-based and machine-learning-based methods are the most commonly used. A method that has been widely applied to detect driving heterogeneity is the fuzzy logic method (Aljaafreh et al., 2012; Alpar and Stojic, 2016; Castignani et al., 2015; Choudhary and Ingole, 2014; Dorr et al., 2014). For example, Aljaafreh et al. (2012) proposed a fuzzy inference system that uses the average of the Euclidean norms of longitudinal acceleration and lateral acceleration to infer driver type. Castignani et al. (2015) presented a profiling and scoring application that can accurately identify risky driving events such that aggressive and calm driving styles can be easily distinguished. Dorr et al. (2014) developed an online system using fuzzy logic to detect driver types in real time.

On the other hand, machine learning methods, including artificial neural networks (ANNs), support vector machines (SVMs), and random forest decision methods, have been widely used to classify driver heterogeneity. Mohammadnazar et al. (2021) proposed a framework that uses basic safety messages generated by connected vehicles to identify driver heterogeneity using unsupervised machine learning methods. Wang et al. (2017) used the K-means and SVM methods to group driver heterogeneity into aggressive and normal styles. Brombacher et al. (2017) used an ANN to quantify and classify driver heterogeneity. They grouped driver style into five categories, namely, "very sporty," "sporty," "normal," "defensive," and "very defensive". Bejani and Ghatee (2018) used a fusion technique to combine SVM, K-nearest neighbors, and multi-layer perception to recognize different types based on smartphone data. Thus, driving heterogeneity detection and prediction of surrounding HDVs should be incorporated into ADS motion planning.

## 2.2 Trajectory prediction of surrounding HDVs

The trajectory prediction of surrounding HDVs has attracted considerable interest over the past decade. Trajectory-prediction models are used to obtain the future motion of surrounding HDVs in advance to enable the ADS to make a good decision. Early research on this topic used simplified kinematic models with a constant yaw rate and acceleration to predict the future trajectories of vehicles. However, this approach does not consider multiple maneuvers, particularly highly transient maneuvers, such as lane changes (Kim and Kum, 2018). To address this limitation, researchers have proposed a maneuver-based trajectory-prediction method that estimates both driving maneuvers and future trajectories. One method to achieve maneuver-based trajectory prediction is the use of a maneuver-recognition module, which selects the current maneuver from a predefined set by comparing the centerlines of the road lanes to a local curvilinear model of the vehicle path (Houenou et al., 2013). Kasper et al. (2012) proposed object-oriented

Bayesian networks to address the recognition of driving maneuvers such as overtaking, lane changing, and car following based on training data.

Another approach for trajectory prediction of surrounding HDVs is to consider the traffic context in the motion prediction. This can stabilize the results in complex environments, which is another focus of this study. This approach involves a learning-based method to classify and predict the trajectories of vehicles surrounding an ADS. Vehicles are described as maneuvering entities that can interact with each other (Xing et al., 2020). Previous studies on this can be classified into two categories. The first category focuses on learning mutual dependencies between multiple entities using a probabilistic model. For example, Murphy (2002) presented a probabilistic agent interaction model to infer vehicle behavior. The model provides a natural framework that estimates the agent status values from data and performs statistical inference. The model proposes two dynamic layers: dynamic and context layer. The dynamic state stores low-level information such as position and orientation, whereas the context layer contains high-level semantic information such as the relative locations of surrounding vehicles. The future motion of a vehicle can be inferred using a conditional probability distribution over states. Agamennoni et al. (2012) proposed a Bayesian dynamic network to estimate the state of traffic participants and anticipate their future trajectories, which considers low-level vehicle information such as pose, orientation, and velocity from sensors, and high-level semantic information such as the relative distance between surrounding vehicles. Bahram et al. (2016) presented a game-theoretic algorithm to estimate the motion intentions of surrounding vehicles. This enables the integration of traffic laws when combining maneuver sequences over multiple time steps instead of a single future maneuver. Li et al. (2019) proposed a Bayesian dynamic network with multiple predictive features, including the historical states of predicted vehicles, road structures, and traffic interactions.

The second category of methods for modeling vehicle interactions and predicting the trajectory of surrounding vehicles involves recent machine-learning-based approaches. Deo et al. (2018) presented an LSTM encoder–decoder model for predicting vehicle maneuvers and trajectories. Zhao et al. (2019) proposed an improved social LSTM framework called a multi-agent tensor fusion network for multi-step trajectory prediction based on encoding a scene context and historical vehicle trajectories. Zyner et al. (2020) constructed an RNN to predict driver intent at urban intersections using trajectory prediction with uncertainty. They also proposed a clustering algorithm to produce a ranked set of possible trajectories. To validate the model, they collected real-world trajectory data using a lidar-enabled vehicle instead of an overhead camera. Ju et al. (2020) proposed a multi-layer architecture that uses interaction-aware Kalman neural networks (IaKNNs) to predict the trajectory of surrounding vehicles in a dynamic environment. The primary advantage of learning-based methods is that they can learn parameterized policies and do not require perfect knowledge of a vehicle or its environment. Although most studies have focused on predicting the motion of surrounding vehicles, they have considered only a unified driving style for these vehicles. However, these methods of vehicle motion prediction can only anticipate short-term future trajectories. Traditional RNN or LSTM can predict the trajectory up to 2–6 s in advance with a higher accuracy (Deo and Trivedi, 2018; Kim et al., 2017). Thus, an efficient vehicle trajectory model is required. A transformer can be a suitable approach because of its superior learning ability and accuracy.

The transformer model is a deep learning architecture that aims to process input sequences of variable lengths and generate output sequences. It was developed for natural language processing (NLP) tasks and was introduced by Vaswani et al. in 2017. The self-attention mechanism of the transformer enables a model to selectively consider different parts of the input sequence, which can enable better performance in long-sequence prediction (Vaswani et al., 2017). In addition, transformers have been used in innovative research applications, such as ChatGPT (OpenAI, 2023). However, they are rarely used in transportation field. To the best of our knowledge, it has only been used for vehicle intention prediction (Guo et al., 2022) and short-term trajectory prediction (Liu et al, 2021; Singh and Srivastava, 2022; Gao et al, 2023). This is one of the first studies to utilize transformer technology for long-term vehicle trajectory prediction (i.e., more than 10s prediction horizon) in ADS motion planning.

## 2.3 ADS motion planning

In recent years, researchers have conducted numerous investigations into the motion planning of ADSs. Graph search-based planner methods, such as the Dijkstra and state lattice algorithms, have been explored and can determine paths on graphs. For example, Marchese (2006) proposed a graph-searching algorithm that determined the shortest single-source path on a graph. Howard and Kelly (2007) applied local queries from a set of lattices containing all the feasible features, enabling vehicles to travel from their initial state to several other states. The best path can be determined using the cost function between precomputed lattices. In addition, sampling-based planner methods have been proposed for ADS motion planning. Karaman and Frazzoli (2011) proposed fast online path planning by executing a random search considering non-holonomic constraints such as the maximum turning radius of the vehicle. Computer-aided geometric design technology has recently attracted interest for ADS motion planning. It enables a curve planner to fit a given description of a road, considering the feasibility, comfort, and vehicle dynamics to plan the trajectory. This interpolating curve planner uses a given set of waypoints to generate a new smoother path and re-enter a previously planned path (Gonzalez et al., 2014; Rastelli et al., 2014). However, the current challenges in motion planning include real-time planning calculations in dynamic environments. As mentioned in the introduction, the accurate prediction of the surrounding environments of an ADS is critical for generating a new collision-free trajectory within a limited planning time.

The state of the art indicates that accurately predicting the long-term trajectory of surrounding HDVs for ADS motion planning remains challenging and requires further investigation. This study aimed to address this challenge.

## 3. Problem description

This study considered a three-lane highway segment with an on-ramp located on the right-hand side, as shown in Fig. 2. When HDVs arrive at a freeway entrance, they must perform mandatory lane changes to enter the freeway. Assuming that an HDV merging trajectory follows the curve from point A to point B, the lane-change point is defined as the center of the HDVs crossing the boundary of the current lane and the target lane. The following ADS in the target lane must plan its trajectory based on the trajectory of the merging HDV and surrounding vehicles. In the baseline scenario, we assume that the ADS begins to react to the merging HDV after it crosses the lane boundary (i.e., merging point).

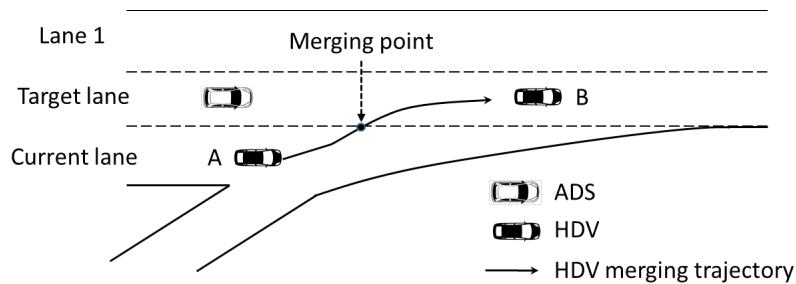

**Fig. 2.** Schematic of HDV merging at the highway segment near an on-ramp

## 4. Methodology

We propose a hierarchical model that has two levels: driving style and trajectory prediction. A driving-style prediction model is designed to identify driving heterogeneity. The transformer model is used to predict the HDV trajectory at the next moment based on data collected over several past moments by leveraging self-attention mechanisms to learn vehicle patterns and dynamics.

### 4.1 Trajectory data pre-processing

In this paper, trajectory data are collected from drones. The data contain the location and speed of the vehicle with a timestamp, including latitude and longitude coordinates $(x, y)$, estimated speed $(v)$ and acceleration $(a)$. The lateral position deviation $(D)$ between the center of the vehicle and the left boundary of the current lane is calculated based on lane markings. For each trajectory data point, the HDV car-following information, such as gap ($G$, the space gap between an HDV and its leading vehicle), and range rate ($RR$, the speed difference between the HDV and its leading vehicle), can be derived directly based on the trajectories of two HDVs. The change rate $K$ of the lateral position deviation is used to predict the driving style when the HDV begins the merging process. The notations for the major trajectory variables used in this paper are shown in Table 1.

Table 1 Notations of major variables used in this paper

| Symbol | Description |
| --- | --- |
| $t$ | Time instant |
| A, B, LCP | The start/end point during the merging process, lane change point |
| $x^h(t), y^h(t), v_y^h(t), v_x^h(t), a_y^h(t),$ | The position, speed (lateral, longitudinal), acceleration of an HDV $h$ at time $t$ |
| $x^a(t), y^a(t), v_y^a(t), v_x^a(t), a_y^a(t),$ | Position, speed (lateral, longitudinal), acceleration of an ADS $a$ at time $t$ |
| LCD | Duration between points A and B |
| $D_t$ | Lateral position deviation at time $t$ between the center of the vehicle to the left boundary of the current lane. |
| $t_A$ | Time when the vehicle is at point A |
| $K_t$ | Change rate of $D_t$ |
| $G_l(t), RR_l(t)$ | Gap and range rate between HDV and leading vehicle at time $t$ on the current lane |
| $G_f(t), RR_f(t)$ | Gap and range rate between HDV and following vehicle at time $t$ on the current lane |

### 4.2 Driving behavior style recognition and prediction model

K-means clustering is used to differentiate driving style groups (i.e., aggressive and normal) of HDVs based on driving behavior characteristics, such as merging duration ($LCD$) and the distance/speed differences between the HDV and surrounding vehicles ($G, RR$). The K-means algorithm partitions all the HDV trajectory data $n$ into $J$ disjoint subsets by minimizing the square error. Let $X = \{x_i, i = 1, ..., n\}$ represent the $n$ $d$-dimensional trajectory data point, $C = \{c_j, j = 1, ..., J\}$ represent the $J$ clusters, and $m_j$ represent the mean of the cluster $c_j$. The square error $S$ between $m_j$ and the trajectory data points in cluster $c_j$ can be expressed as follows.

$$S(c_j) = \sum_{x_i \in c_j} \| x_i - m_j \|^2 \tag{1}$$

$$S(C) = \sum_{j=1}^{J} \sum_{x_i \epsilon c_j} \| x_i - m_j \|^2 \tag{2}$$

The K-means algorithm has three steps: (1) the mean of cluster $c_j$ ($m_j$) is initialized randomly, and the object variables in the dataset are partitioned into the nearest cluster; (2) the new centers of the clusters are reformed, and new clusters are repartitioned based on the new centers; (3) iteration is continued until the cluster assignments stop changing.

HDV trajectory data are grouped into two types based on the above algorithm: label 0 for aggressive driving behaviors and label 1 for normal driving behaviors. Some indicators obtained from the clustering results (such as the change rate $K_t$ of the lateral position deviation $D_t$ at time $t$ between the center of the vehicle and the left boundary of the current lane) are then used to predict the driving style when the HDVs begin the merging process. Logistic regression can be used to predict the likelihood of a driving style. The probability of each driving style is computed based on the change rates over several past moments. $K_t(f)$ is defined as follows:

$$K_t(f) = \frac{D_t(f) - D_A(f)}{t - t_A} \tag{3}$$

where $K_t(f)$ is the rate of change $D_t$ at time $t$ for HDV $f$ ($f = 1, 2, ..., N$, $N$ is the total number of HDVs). $D_A$ is the lateral position deviation of vehicle $f$ between the center of the vehicle and the left boundary of the current lane of the HDV at point A. $t_A$ is the time instant when the HDV is at point A (begins the merging process). The logistic function is defined as follows:

$$y' = \frac{1}{1 + \exp(-z)} \tag{4}$$

$$z = b + \beta_1 K_1(f) + \beta_2 K_2(f) + \cdots + \beta_t K_t(f) \tag{5}$$

where $y'$ is the output of the logistic regression model, $\beta$ is the learned weight, and $b$ is the bias. Training and test sets were used to test the accuracy of the model.

**4.3 HDVs merging trajectory prediction**

We formulate the prediction of an HDV merging trajectory as a regression task. Specifically, the HDV-merging trajectory is considered as time-series datum from the time step ($t_0$) to ($t_T$), whose length ($T$) represents the merging duration (Fig. 3). The value of the trajectory point at the next moment can be calculated based on the data collected over several past moments. The sliding window approach is adopted to merge the trajectory prediction in the next time step, and a state-of-the-art transformer model is proposed to predict the value of the upcoming trajectories. Given that the HDV merging trajectory contains data points $X_{t_0}, X_t, \ldots, X_{t_T}$, the input $X$ of the transformer model is $X_{t_0}, \ldots, X_{t_w}$, and the output $Y$ is $X_{t_w}, X_{t_w+1}, \ldots, X_T$. Each trajectory data point can be a vector containing multiple features, i.e., the locations and velocities of the vehicles.

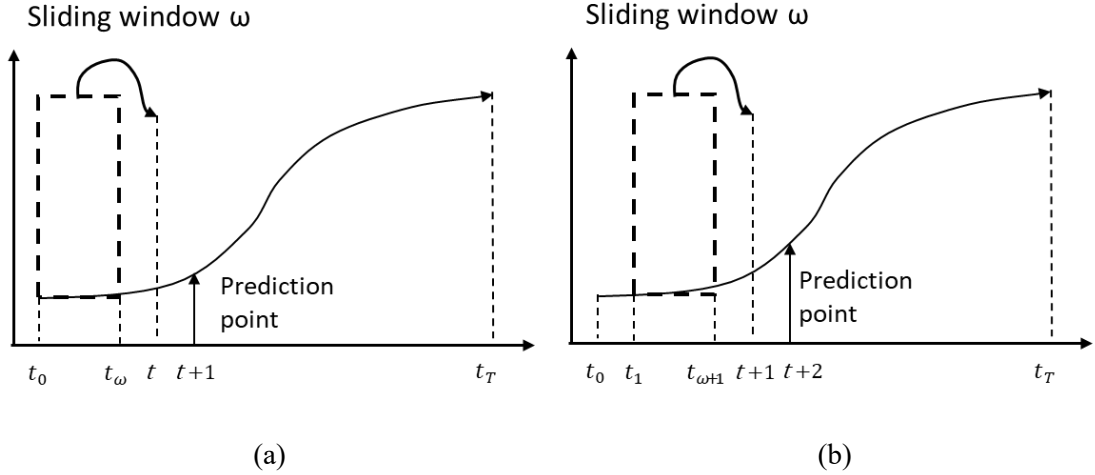

**Fig. 3**. Sliding windows schematic: (a) generated prediction at time step $t + 1$; (b) generated prediction at time step $t + 2$

### 4.3.1 Sliding window approach

In this study, a fixed length sliding time window approach is adopted from Kumar et al. (2013) to use the trajectory data collection during prior time steps to predict the data point in the next time step. In Fig. 3 (a), a window of fixed size $w$ is created to collect the input data. At time step $t$, the model uses the data collected between $t - t_w$ and $t$ to predict the value of the trajectory point at time timestep $t +1$, as shown in Fig. 3(b). The generated prediction is labeled from $t$ and end at $t_T$. The sliding-step duration is 0.033 s. The prediction horizon can be computed using the time difference between $t_T$-$t_0$ and $t - t_0$, $t_p$ is the start point of prediction.

### 4.3.2 Transformer-based trajectory prediction model

The trajectory-prediction model is adopted from the transformer model, which is a deep learning model that uses the self-attention mechanism to learn dependencies in a sequence by processing sequential input data. For a detailed description of the transformer model, please refer to Vaswani et al. (2017). As shown in Fig. 4, the transformer model has an encoder–decoder architecture. The encoder section contains three layers: an input layer, a positional encoding layer, and a stack of four identical encoder layers. The input trajectory data at each time point are converted by embedding them into a vector. The positional encoding layer then aims to encode sequential information through element-wise addition of the input vector. Each encoder layer generates encodings containing information regarding the parts of the inputs that are relevant to each other. Two sub-layers are included in each encoder layer: self-attention and fully connected feedforward. The decoder design is similar to that of the encode block, which contains an input layer, four identical decoder layers, and an output layer. Both encoder and decoder layers use an attention mechanism. In contrast to the encoder layer, each decoder layer uses an additional attention mechanism to obtain relevant information from the outputs of the previous decoders. The decoder also inserts a sub-layer to utilize the self-attention mechanism in the encoder output. Finally, the output layer of the decoder maps the output of the last decoder layer to predict the time sequence.

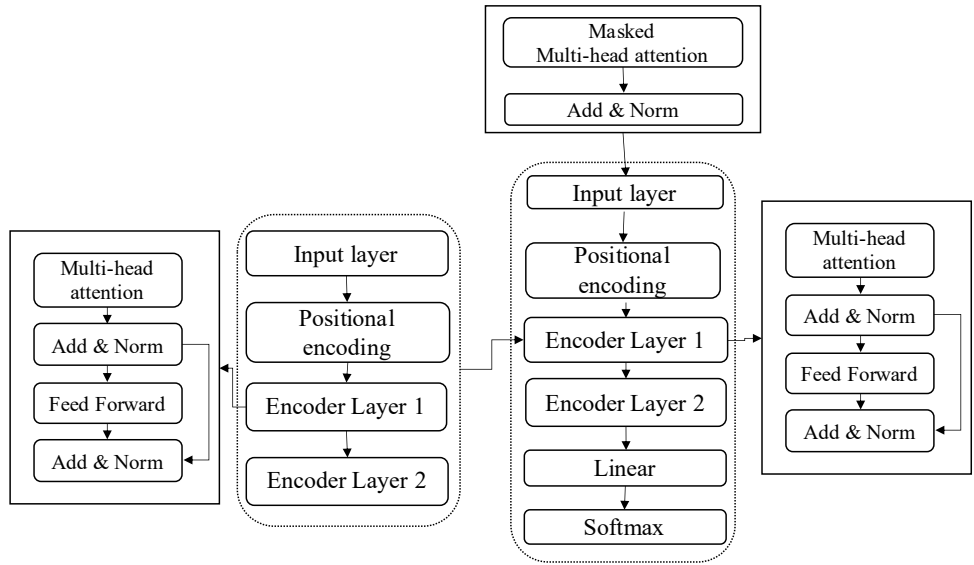

**Fig. 4**. Architecture of the transformer model

Positional encoding is proposed to aid the model incorporate the order of words because the transformer model is a non-recurrent architecture of multi-head attention. Absolute positional encoding based on the sine and cosine functions is expressed as follows:

$$PE_{(pos,2i)} = \sin(pos/10000^{2i/d_{model}}) \tag{6}$$

$$PE_{(pos,2i+1)} = \cos(pos/10000^{2i/d_{model}}) \tag{7}$$

The transformer model uses an attention function that maps a query and a set of key-value pairs to an output, as shown in Fig. 5(a). A query is a feature vector representing the aspects that should be attended to in a sequence. The keys and values are the feature vectors. The score function $f_{attn}$ is used to rate the elements by taking the query and key as the input and the score/attention weight of the query-key pair. The weights of the average $\alpha_i$ and output are calculated using a softmax overall score function output as follows:

$$\alpha_i = \frac{\exp(f_{attn}(key_i, query))}{\sum_j \exp(f_{attn}(key_i, query))} \tag{8}$$

$$output = \sum_i \alpha_i \cdot value_i \tag{9}$$

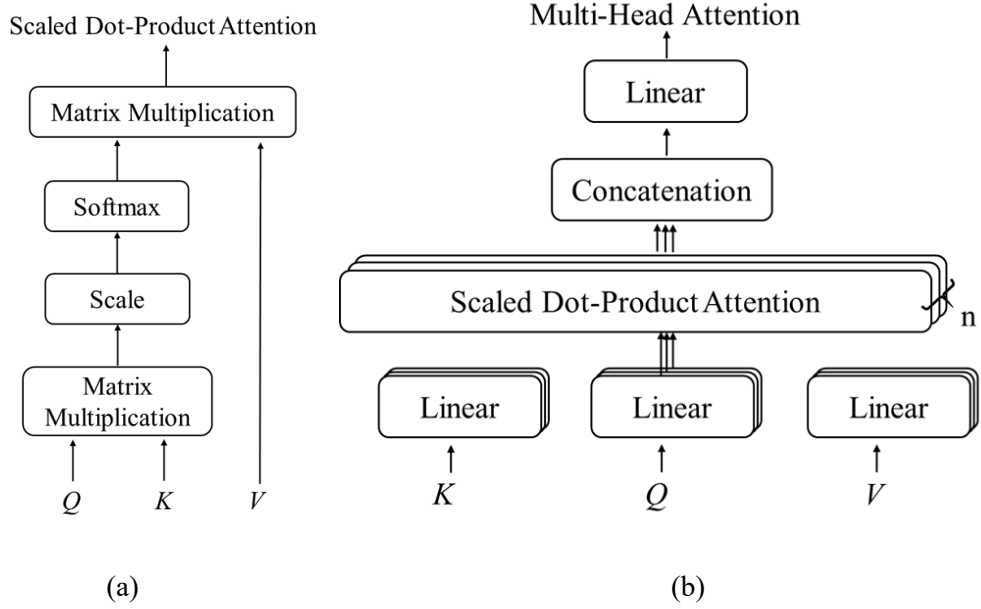

**Fig. 5.** Attention mechanism in a transformer.

The core concept of attention in the transformer model is scaled dot-product attention. A set of queries $Q \in R^{T \times d_k}$, keys $K \in R^{T \times d_k}$, values $V \in R^{T \times d_k}$, where $T$ is the sequence length, and $d_k$ is the hidden dimensionality of queries/keys, and $d_v$ is the hidden dimensionality of values. Dot production attention uses queries and keys as inputs. The attention value from trajectory points $i$ to $j$ is estimated using the dot product attention as follows:

$$Attention(Q, K, V) = softmax\left(\frac{QK^T}{\sqrt{d_K}}\right)V \tag{10}$$

where the matrix multiplication $QK^T$ is the dot product for each pair of queries and keys divided by $\sqrt{d_K}$, and a softmax function is applied to obtain the weights of the values.

The attention mechanisms are extended to multiple query key–value triplets on the same features in the transformer model, which can be concatenated and combined with a final weight matrix, as shown in Fig. 5(b). Specifically, a query, key, and value matrix are transformed into $h$ sub-queries, subkeys, and sub-values. Mathematically, this operation can be expressed as follows.

$$Multihead(Q, K, V) = concat(head_1, \dots, head_h)W^O \quad (11)$$

$$head_i = Attention(QW_i^Q, KW_i^K, VW_i^V) \quad (12)$$

where the learnable parameters $W_{1\dots h}^Q \in R^{D \times d_k}$, $W_{1\dots h}^K \in R^{D \times d_k}$, $W_{1\dots h}^V \in R^{D \times d_v}$, and $W_{1\dots h}^O \in R^{h \cdot d_k \times d_{out}}$. The input trajectory vector $X$ can be transformed into the corresponding feature vectors (queries, keys, and values) using the consecutive weight matrices $W^Q$, $W^K$, and $W^V$.

## 4.4 Improving ADS safety strategy based on prediction

Given the predicted merging trajectory of the HDV, the ADS can incorporate the results into its trajectory-planning module to improve safety. Four driving states are adopted from existing literature: (1) cruising, (2) following, (3) real-time avoidance (RTA), and (4) avoidance in advance (AIA) (Huang et al., 2020; Marino et al., 2011). The four driving states are used to determine the ADS vehicle state, as shown in Fig. 6. $\Delta D(t)$ is the longitudinal distance difference at the timestep $t$ between the HDV and ADS. The HDV is in the cruising state after it merges if $\Delta D(t)$ is longer than the cruising distance $D_c$ (Fig. 6(a)). The ADS enters the following state if $\Delta D(t)$ is within the distance $D_f$, which is between $D_s$ and $D_c$ (Fig. 6(b)). If $\Delta D(t)$ is smaller than $D_s$, then the HDV and ADS are in the unsafe zone, the ADS will enter the real-time avoidance state after HDV merges if the ADS does not predict the HDV trajectories in advance. This state occurs when the HDV merges into an unsafe zone and a high deceleration rate is required to avoid collisions (Fig. 6(c)). If the HDV trajectories are predicted in advance, and the predicted HDV enters the unsafe zone, the ADS performs collision avoidance (i.e., decelerates) in advance before the HDV merges into the target lane (Fig. 6(d)). Dotted parts of the HDV trajectory represent the predicted HDV trajectory. Based on the state of the ADS, reference velocity $v_{ADS}(t)$ can be calculated as follows:

(1) When $\Delta D(t) > D_C(t)$, then ADS will enter the cruising state:
$$D_C(t) = D_{C0} + \alpha v_{ADS}(t-1) \quad (13)$$
$$v_{ADS}(t) = v_C \quad (14)$$

where $v_C$ is cruising velocity, $v_{ADS}(t_0) = 13.7\ m/s$.

(2) When $D_s(t) < \Delta D(t) < D_C(t)$, then ADS will enter the following state:

$$D_S(t) = D_{S0} + \partial v_{ADS}(t-1) \tag{15}$$
$$D_f(t) = D_{f0} + \gamma v_{ADS}(t-1) \tag{16}$$

$$v_{ADS}(t) = v_{HDV}(t) + \delta(\Delta D(t) - D_f(t)) \tag{17}$$

where $v_{ADS}(t)$ denotes the reference velocity of the ADS at time $t$, $v_{HDV}(t)$ denotes the velocity of the vehicle ahead at time $t$, $D_f$ is the following distance.

(3) When $\Delta D(t) < D_s(t)$, and the ADS is in the RTA state:

$$v_{ADS}(t) = v_{0,\ RTA}(t_0) - \int_{t_0}^{t} a_d(t) dt \tag{18}$$

where $v_{0,\ RTA}(t_0)$ is the initial velocity of the ADS when the ADS begins RTA. $a_d(t)$ is the maximum deceleration.

(4) When $\Delta D(t) < D_s(t)$, the ADS is in the AIA state with the predicted HDV trajectory:

$$\int_{t_{0,\ AIA}}^{t_{0,\ AIA}+t_p} [v_{HDV}(t) - v_{ADS}(t)] dt = D_f(t) - D_{0,\ AIA}(t_0) \tag{19}$$

where $D_{0,\ AIA}(t_0)$ are the initial distances between the HDV and ADS when the ADS enters the AIA state.

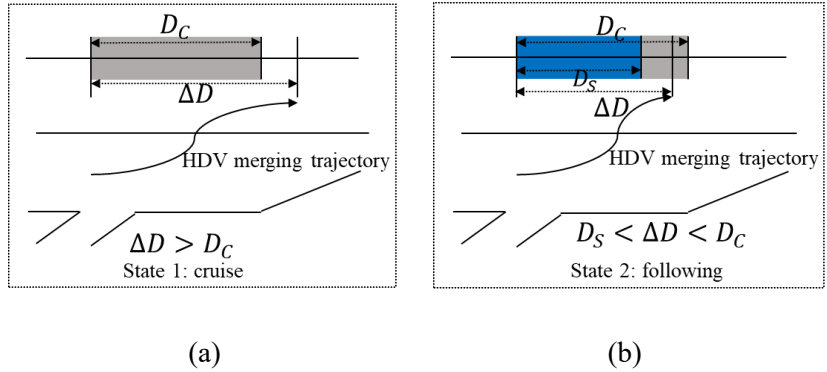

(a) State 1: cruise, $\Delta D > D_C$

(b) State 2: following, $D_S < \Delta D < D_C$

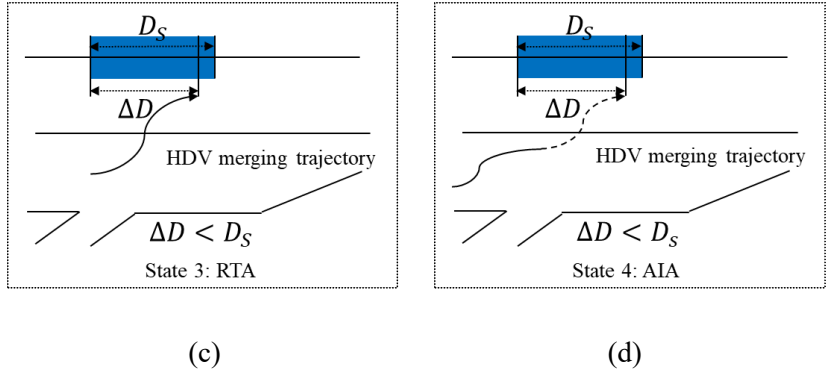

(c)                  (d)

**Fig. 6.** Interaction between an HDV and ADS: (a) State 1: cruise; (b) State 2: following; (c) State 3: real-time avoidance; (d) State 4: avoidance in advance.

## 5 Case study

In this section, an actual merging scenario is presented to verify the proposed methods for ADS trajectory planning. First, the HDV-merging behavior trajectory data were processed from video data. Second, prediction methods based on logistic regression and transformers were trained and tested to evaluate their performance. Finally, decision-making strategies based on the prediction results were applied to highway merging scenes.

### 5.1 HDVs merging trajectory processing

Field data were collected in the merging area of the Qujiang interchange in Xi'an, China (Fig. 7(a)). The video data were collected using a small-scale quadcopter camera drone. The data collection and processing consisted of four steps. Step 1: Trajectory data were detected and tracked from the UAV video using a computer-version technique. In Fig. 7(a), the yellow rectangles represent the bounding boxes of the tracked vehicles. Subsequently, the vehicle information in each video frame, including the lateral coordinate $x$, longitudinal coordinate $y$, vehicle ID $id$, frame ID $f$, length of the vehicle in feet $l$, and width of the vehicle in feet $w$, was be extracted. Further car-following behavior and vehicle dynamics data, such as range ($R$, the space difference between the HDV and its leading vehicle), range rate ($RR$, the speed difference between the HDV and its leading vehicle), and velocity $v$ and acceleration $a$ of the HDV, were obtained. Step 2: A line boundary and lane number identification algorithm was applied. The recognition area used in each frame of the video was designated using the OpenCV tool, as shown in Fig. 7 (b). The coordinates of each lane were obtained. Based on the position of each vehicle in each frame, a clustering algorithm was used to determine the lane number of each vehicle. Step 3: A merging trajectory extraction algorithm was applied to obtain specific merging scenarios. This algorithm tracked the lane number assigned to each vehicle and identified the lane-changing moment when the vehicles crossed the road line. The trajectory of the lane-changing vehicle was then extracted and divided into two parts: pre-lane changing and during lane changing. Some important timestamps were defined in the lane-changing trajectory, including the start moment of the lane-change process $t_S$, end moment of the lane-change process $t_S$, lane-change moment $t_{lc}$, and pre-lane-change moment $t_{pre}$. Step 4: The information of surrounding vehicles was extracted. The vehicles leading and following the merging vehicle at the start of the lane change were defined as the surrounding vehicles. The distance and speed difference between the surrounding and merging vehicles were calculated.

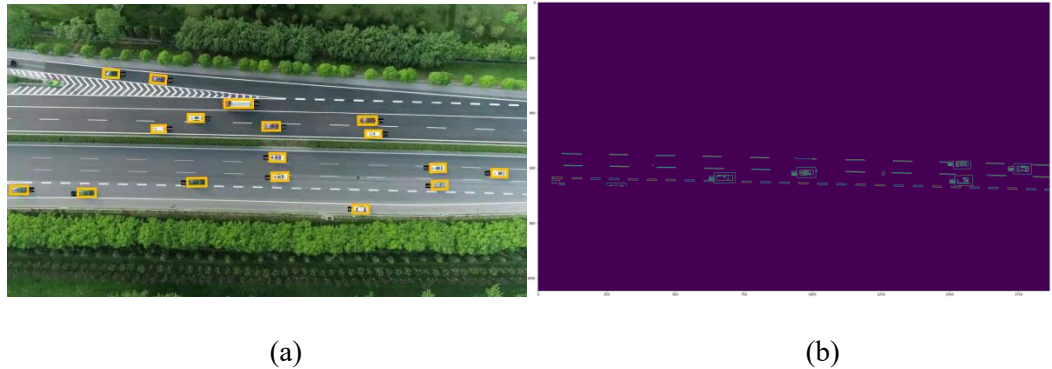

(a)                                                                                  (b)

**Fig. 7.** (a) Vehicle trajectory merging area of Qujiang interchange in Xi'an, China. (b) Recognition area in OpenCV

## 5.2 Prediction of HDV merging style

Trajectory data were used to train and test the proposed prediction method. To obtain data with different merging styles of the HDVs, we classified the trajectory data using the K-means clustering approach. A total of 202 merging HDV samples were extracted. The sample sizes of the two merging behavior groups (aggressive and normal) were 99 and 103, respectively. Based on the classification results, the change rate of the lateral position deviation, $D_t$, was selected as the input feature to predict the driving style using logistic regression:

$$K_T(f) = [K_1(f), \dots, K_i(f), \dots, K_{lc}(f)] \quad (20)$$

where $K_i(f)$ is the rate of change in the lateral position deviation between the center of the vehicle and the left boundary of the current lane at $i$th time step.

Among the data, 80% were used for training and 20% were used for testing. A confusion matrix was used to measure the prediction accuracy, which showed the predicted value based on the model versus the actual values in the test dataset. It contained four types of results: true positive (TP), false negative (FN), false positive (FP), and true negative (TN). Accuracy can be defined as follows:

$$A_i = \frac{TP+TN}{P+N} = \frac{TP+TN}{TP+TN+FP+FN} \tag{21}$$

where $A_i$ denotes the prediction accuracy at the $i$th time step. The results showed that the highest accuracy $A_i$ was achieved at the fourth time step, which meant that most of the HDV driving styles could be predicted correctly ($A_i = 0.74$) after 0.13 s.

### 5.3 Prediction of HDV merging trajectory

A trajectory dataset from the UAV was used to train and test the merging trajectory-prediction method. The trajectory dataset for merging behavior included the lateral and longitudinal coordinates, velocity, acceleration, and the current lane position of HDVs. The sample sizes of the two merging behavior groups (aggressive and normal) were 99 and 103, respectively. 70% of the data were used for model training and the remaining 30% were used for model testing. The samples for the two groups were processed using the sliding window approach to generate training and testing data, as shown in Fig. 3. The prediction time sequence was considered the input for the proposed transformer-based prediction model (Section 4.2). The transformer prediction method was used, which included six layers: (1) $4 \times \omega$-neuron input layer, (2) 12-head multi-head attention layer, (3) 64-neuron feed forward network, (4) global average pooling 1D layer with 10% dropout, (5) 64-neuron dense layer with ReLU activation function and 10% dropout, and (6) 4-neuron output layer with SoftMax activation function. The loss function was the mean squared error (MSE) and the optimizer used was RMSRrop. The input matrix of the model collected from time step $t - t_w$ to $t$ was used to predict the value of the trajectory point at time timestep $t +1$ in the sub-window as follows:

$$X_i = (x^h, y^h, v_x^h, v^h) \tag{22}$$

where $X_i$ describes the lateral coordinate, longitudinal coordinate, and velocity of the HDV at the $i$th timestep. Different time windows ($w$ = 1.67, 3.33, 5, 6.67, 8.33, 10 and 11.67 seconds) were trained and tested on the transformer model. The MSE was used to assess the overall prediction accuracy of the transformer model; it was calculated as follows:

$$\text{Mean squared error} = \frac{1}{N}\sum_{j=1}^{N}(Y_j - \widehat{Y_j}) \tag{23}$$

where $N$ is the number of vehicles tested, $Y_j$ is the actual value, and $\widehat{Y_j}$ represents the predicted value. As listed in Table 2, the test accuracy could be improved by selecting the appropriate $w$. The best performance with a lower MSE was obtained when the time window $w$ were 6.67 s (normal) and 11.67 s (aggressive), as shown in Table 2. Taking two vehicles (one for normal driving and the other for aggressive driving) as examples, a comparison of the HDVs behavior characteristics is shown in Fig. 8. Aggressive driving exhibited more velocity fluctuations than normal driving. Before the lane change point, normal driving accelerated until a maximum value occurred when the HDV crossed the lane line, and then decelerated sharply. However, during aggressive driving, the driver frequently adjusted the vehicle during the merging process. The velocity increased after the HDV crossed the lane. Significant decelerations in aggressive driving appeared at 13–16 s and 31–36 s, as shown in Fig. 8(b).

Table 2 Performance of the transformer model

| Time windows (seconds) | 1.67 | 3.33 | 5 | 6.67 | 8.33 | 10 | 11.67 |
| --- | --- | --- | --- | --- | --- | --- | --- |
| Normal (MSE) | 0.0534 | 0.0641 | 0.0418 | **0.0314** | 0.039 | 0.0315 | 0.0343 |
| Aggressive (MSE) | 0.047 | 0.049 | 0.031 | 0.029 | 0.0288 | 0.0303 | **0.0280** |

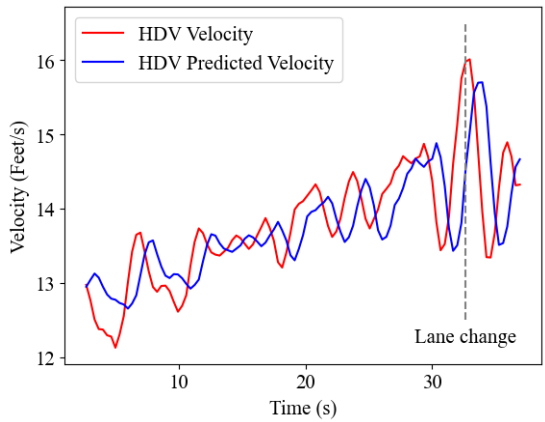
(a)

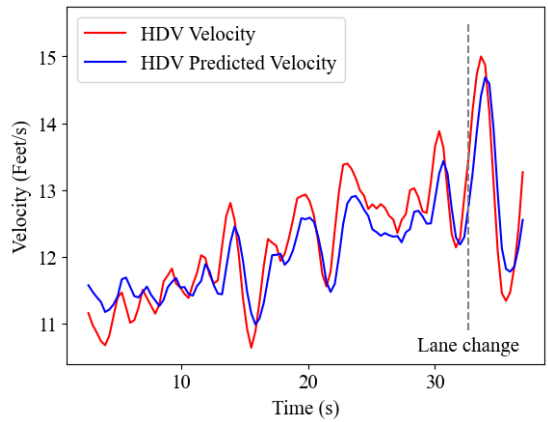
(b)

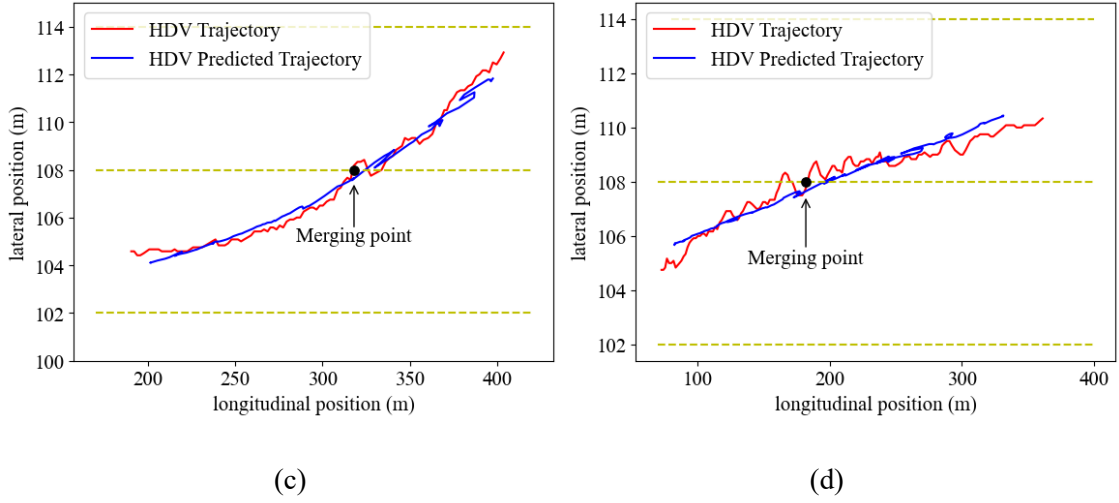

**Fig. 8.** Real and predicted velocities of HDVs: (a) normal style, (b) aggressive style; real and predicted trajectories of HDVs: (c) normal style, (d) aggressive style.

### 5.4 ADS decision-making strategy

Four scenarios were constructed to test the prediction results and proposed strategies. The HDV predicted trajectory and proposed strategies were applied to these scenarios. This interaction occurred after the HDV merged into the target lane. The parameters used in the ADS decision-making strategy are listed in Table 4.

### 5.4.1 Prediction horizon

An ADS trajectory with different prediction horizons can be planned based on the ADS safety strategies when the HDV interacts with the ADS (Section 4.3). Before determining the ADS strategies, the prediction horizon of the HDV must be selected. In this study, the prediction horizon was defined as how early the HDV merging trajectory can be predicted before merging to develop ADS strategies. The TTC between the HDV and ADS was used to evaluate the safety performance of the different prediction horizons. The TTC for the ADS at time step $t$ with respect to the HDV was calculated as follows:

$$TTC_{ADS}(t) = \frac{X_{HDV}(t) - X_{ADS}(t) - L_{HDV}}{V_{ADS}(t) - V_{HDV}(t)} \tag{24}$$

where $X_{HDV}$ and $V_{HDV}$ denote the position and velocity of HDV, respectively. $X_{ADS}$ and $V_{ADS}$ refer to the position and velocity of the ADS, respectively. $L_{HDV}$ is the length of the HDV. Table 3 presents the TTC of the ADS trajectory planning with different prediction horizons. The best performance of the ADS trajectory was obtained when the prediction horizon was 10 s for both the aggressive and normal driving styles. In the following section, we consider vehicle No. 38 (aggressive driving) and No. 55 (normal driving) as examples to analyze the planned ADS trajectory. As shown in Table 3, the safety performance of the ADS planned trajectory with HDV prediction increased compared with that of the ADS trajectory without HDV prediction for both aggressive and normal driving (i.e., the TTC increased from 90.49 to 300.4 and from 101.69 to 383.88, respectively). Compared with aggressive driving, the safety performance of the ADS exhibited a more significant improvement when the trajectory of the ADS could be planned with the HDV-predicted merging trajectory for the normal driving style, as shown in Table 3. This indicated that the predicted HDV trajectory is more effective when the ADS encounters the merging of aggressive HDV.

Table 3 Safety performance of ADS planned trajectory with different predictions.

| Prediction horizon (seconds) | TTC of ADS | |
| --- | --- | --- |
| | Normal driving | Aggressive driving |
| Without prediction | 101.6988 | 90.4932 |
| 3.33 | 209.0554 | 151.3918 |
| 6.67 | 373.1847 | 278.9912 |
| 10 | **383.8816** | **300.4004** |
| 13.33 | 342.8960 | 298.7944 |
| 16.67 | 340.0593 | 300.0429 |
| 20 | 338.4308 | 289.9127 |
| 23.33 | 337.2220 | 267.4048 |
| 26.67 | 336.4743 | 272.5936 |
| 30 | 334.6756 | 280.0317 |

### 5.4.2 ADS motion planning

Table 4 Parameters value of the best strategy

| Parameters | |
| --- | --- |
| $D_{c0}$ | 30 (m) |
| $D_{f0}$ | 9 (m) |
| $\alpha$ | 2 |
| $\partial$ | 0.7 |
| $v_{ADS}(t_0)$ | 13.72 (m/s) |
| $\delta$ | 0.4 |
| $\gamma$ | 0.3 |

Scenario 1: The trajectory of the ADS can be planned based on the longitudinal distance between the ADS and front of the HDV. For this scenario, we selected HDV from the aggressive driving style group. Fig. 9 shows the results of the ADS using the RTA strategy. The longitudinal positions of the vehicles are shown in Fig. 9(a). At 33 s, the HDV merged with the target lane and became the front vehicle of the ADS. The longitudinal distance between the two vehicles became 0.41 m, and the ADS had to reach the RTA state to maintain a safe distance from the HDV until it reached $D_s$ at 34.2 s, as shown in Fig. 9(b). After 34.2 s, the gap was between $D_c$ and $D_s$, which caused the ADS to apply the following state. The velocities of the ADS and HDV are shown in Fig. 9(c). The actual velocity of the ADS was controlled based on the reference velocity. A significant deceleration of the HDV occurred at 31–32.5 s and 38–39 s. After 34.2 s, the ADS accelerated and operated in the following state, and the gap between the HDV and ADS increased.

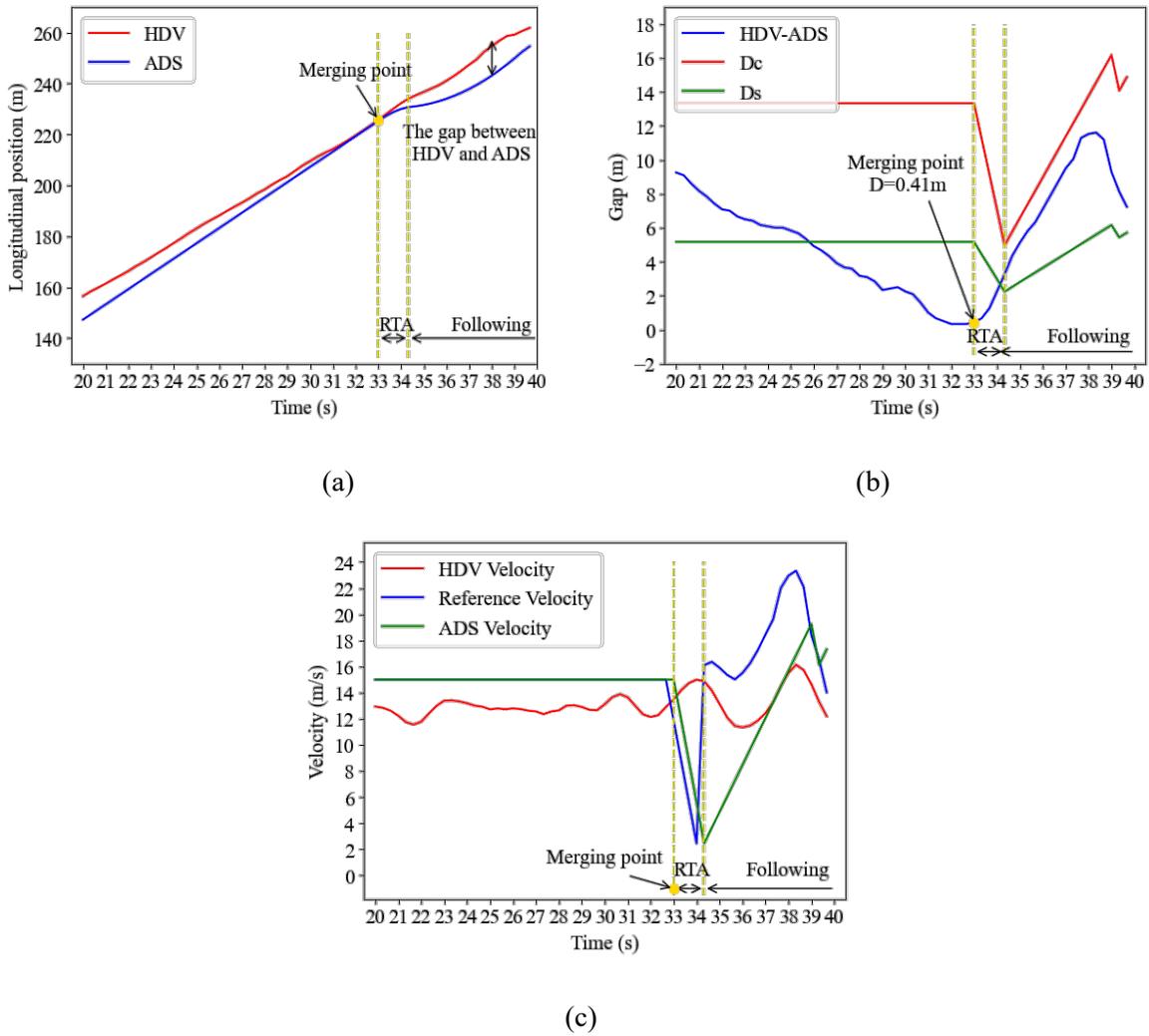

**Fig. 9.** Results of ADS trajectory planning without HDV (aggressive driving) trajectory prediction. (a) Longitudinal position of vehicles. (b) Gap between vehicles. (c) Velocities of vehicles.

Scenario 2: The trajectory of an HDV with an aggressive driving style is predicted before merging into the target lane. Based on the prediction results, the ADS performs collision avoidance (i.e., it decelerates) in advance (AIA) before merging into the target lane. Fig. 10 shows the results of the ADS implementation of the AIA strategy. As shown in Fig. 10(a), the same gap in the longitudinal position between the HDV and ADS was maintained with AIA strategy implementation. At 23 s, the trajectory of the HDV was predicted. The gap between ADS and HDV was predicted to be less than that of $D_s$ after merging into the target lane. Subsequently, the ADS had to implement AIA at 30.7 s until the gap reached $D_s$ at 32 s, as shown in Fig. 10(b). Compared with scenario 1, the distance between the HDV and ADS at the merging point was larger, which indicated that the AIA aided the ADS in maintaining a safe distance from the leading HDV. Compared with the results of the implemented RTA strategy, the gap fluctuation between the two vehicles was smaller, as shown in Fig. 10(b).

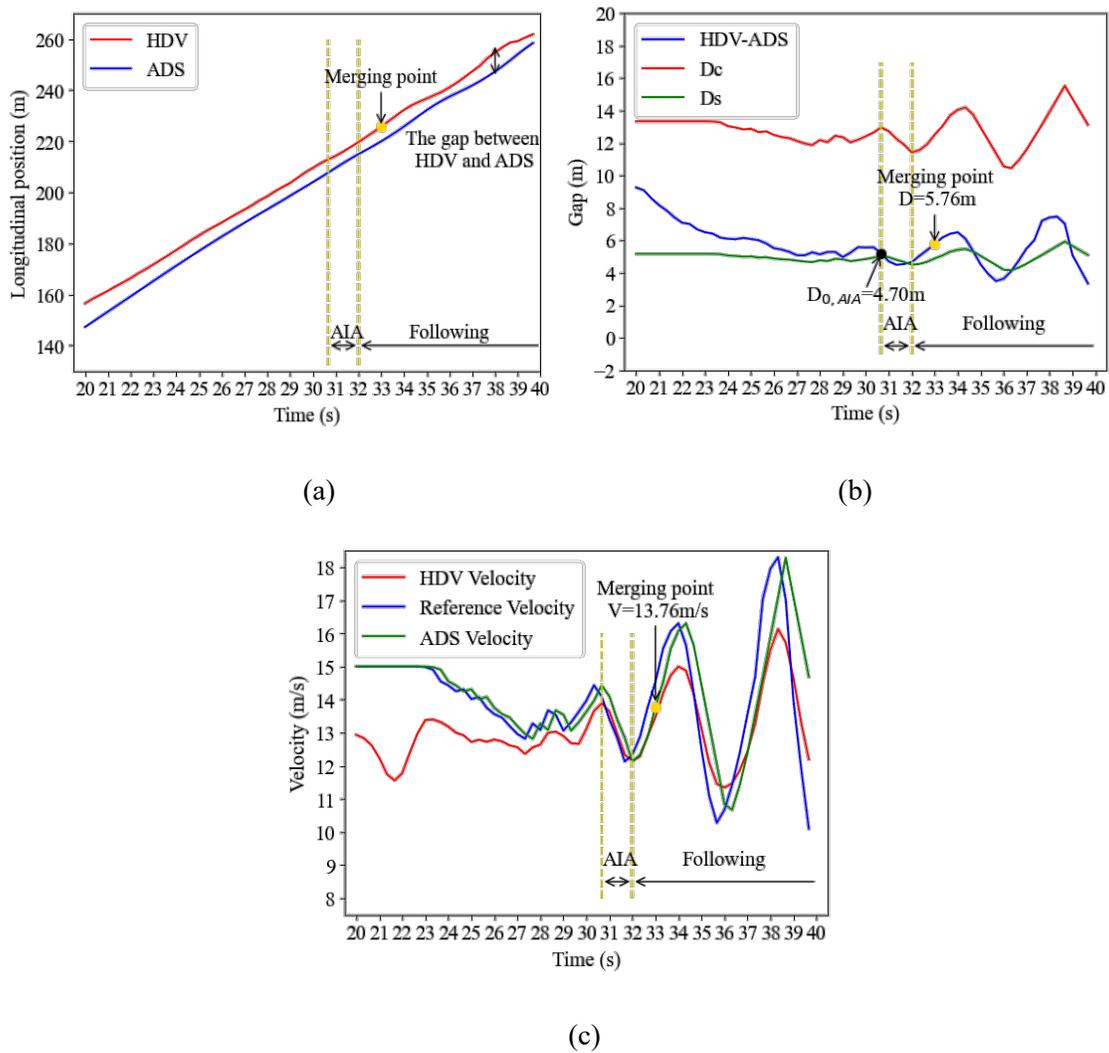

**Fig. 10.** Results of ADS trajectory planning without HDV (aggressive driving) trajectory prediction. (a) Longitudinal position of vehicles. (b) Gap between vehicles. (c) Velocities of vehicles.

Scenario 3: The ADS trajectory is determined based on $\Delta D(t)$. In this scenario, vehicles from the normal-driving group were identified as research targets. As shown in Fig. 11 (a), at 33 s, the HDV merged into the target lane, and the ADS entered the RTA state. The differences in the longitudinal positions between the HDV and ADS increased owing to the sudden deceleration of the HDV. The gap at the merging point was 1.11 m, which was larger than the 0.41 m obtained in the aggressive driving group. This indicated that at the merging point, normal driving is safer than aggressive driving, which is consistent with the previous studies. The reference and actual velocities of the ADS are presented in Fig. 11 (c). At the merging point, the HDV and ADS had almost the same velocity. The deceleration of the HDV had a time delay compared with that of the ADS. Compared with the results of the aggressive driving style, the significant decelerations of HDV with normal driving style after merging increased the gap between vehicles to improve safety performance, as shown in Figs. 9 and 11.

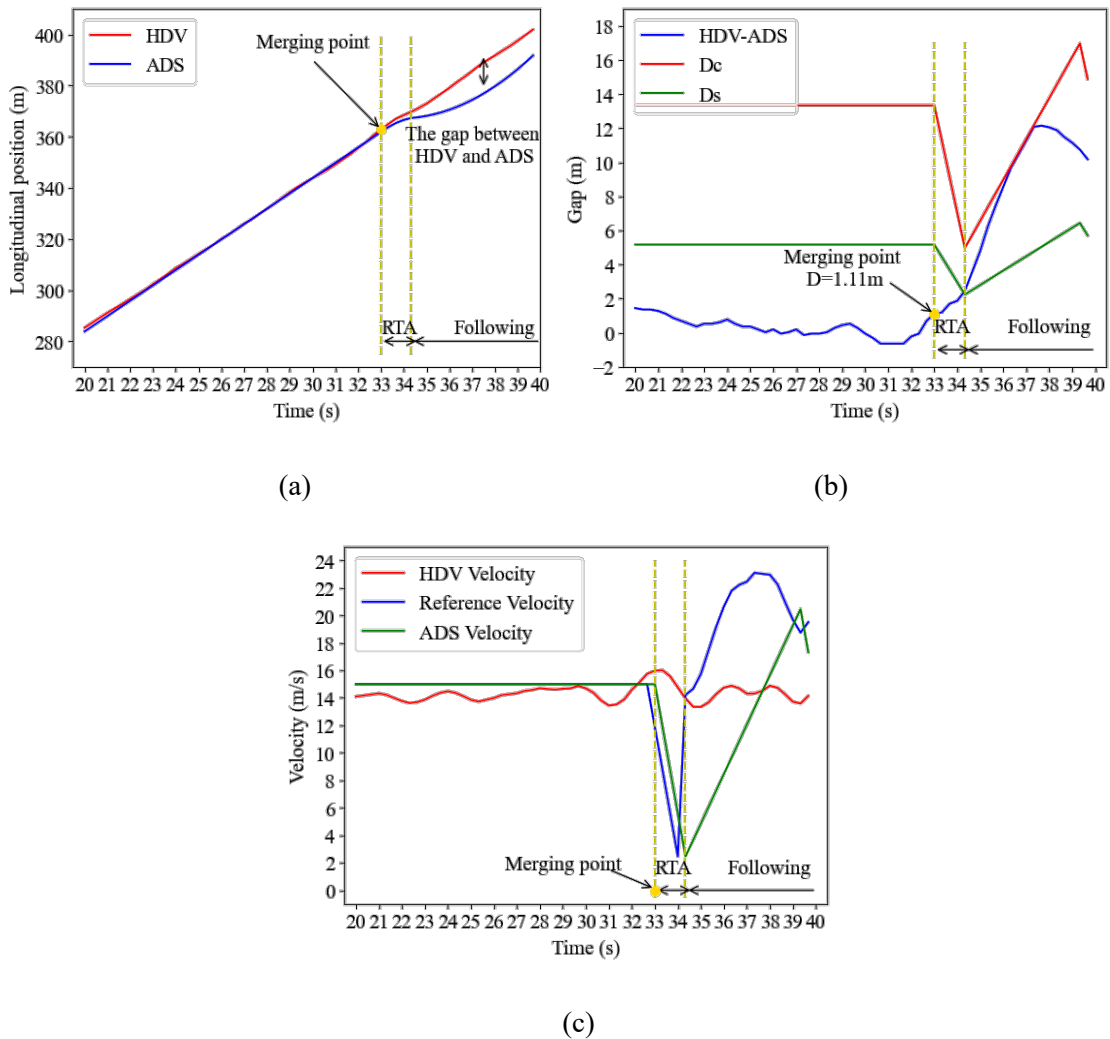

**Fig. 11.** Results of ADS trajectory planning without HDV (normal driving) trajectory prediction. (a) Longitudinal position of vehicles. (b) Gap between vehicles. (c) Velocities of vehicles.

Scenario 4: The trajectory planning of the ADS using the proposed strategy is illustrated in Fig. 12. This strategy can be determined based on the predicted HDV with a normal driving style. At the merging point, the gap between the ADS and HDV is greater than the safe distance because the ADS implements the AIA strategy based on the predicted HDV trajectory, as shown in Fig. 12 (b). Compared with the results of the ADS, which suddenly decelerated the velocity after merging, the AIA avoided a sharp fluctuation in the velocity between 33 and 34.3 s, as shown in Fig. 12 (c). As shown in Fig. 10 (c) and 12 (c), at the merging point, the velocity of the HDV with a normal driving style was greater than that of the HDV with an aggressive driving style when the ADS implemented the AIA strategy. The velocity of the ADS fluctuated more when interacting with an aggressive driving vehicle than when encountering a normal driving vehicle.

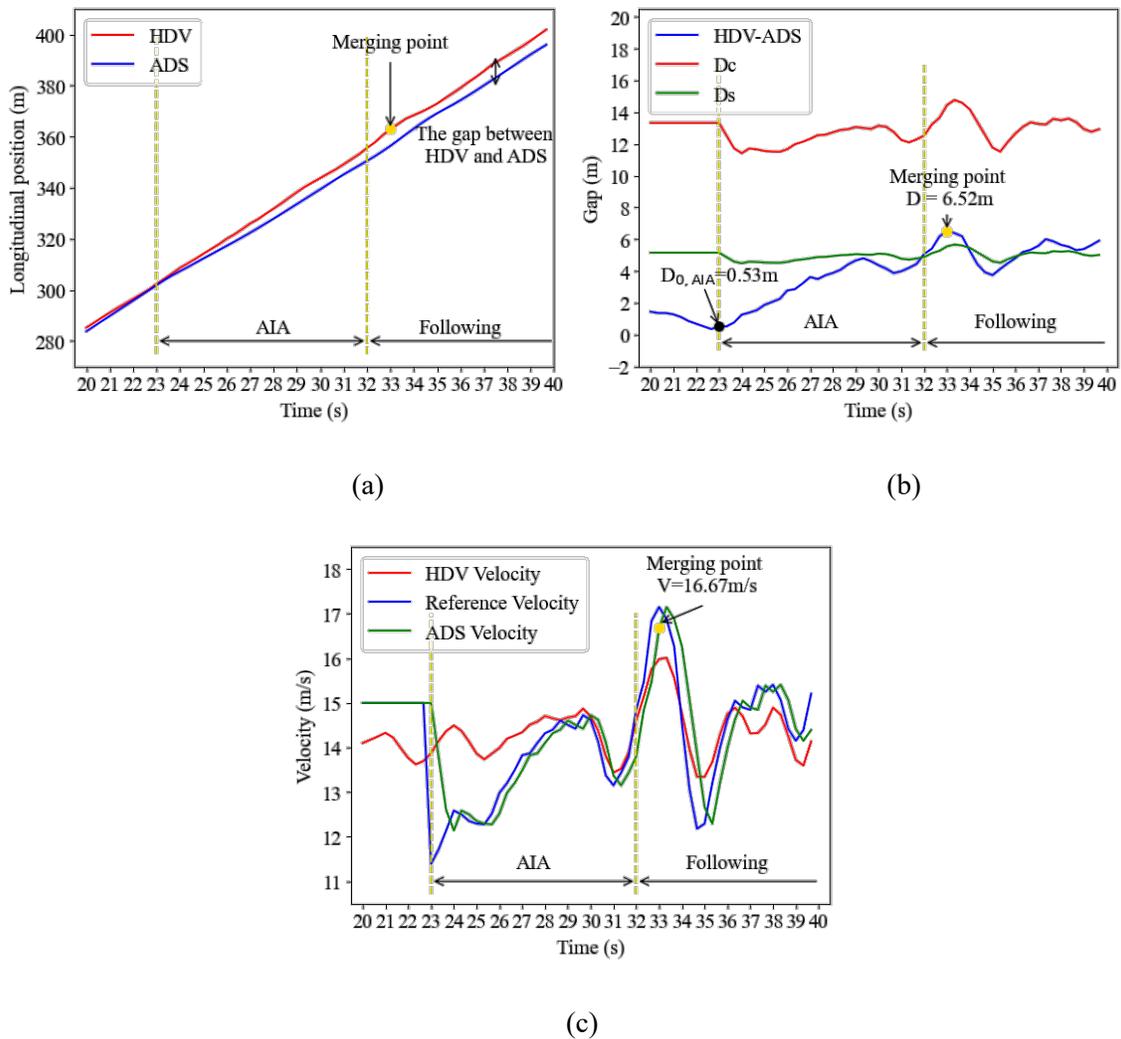

**Fig. 12.** Results of ADS trajectory planning with HDV (normal driving) trajectory prediction. (a) Longitudinal position of vehicles. (b) Gap between vehicles. (c) Velocities of vehicles.

## 6. Conclusion

In this paper, we propose a hierarchical model that combines the results of the driving styles of surrounding HDVs and long-term trajectory for the motion planning of an ADS to improve driving safety. In particular, we apply an unsupervised learning method to group HDVs with similar driving behaviors in a highway ramp-merging scenario. Using these classification results, logistic regression is used to predict the driving style (normal or aggressive driving) based on the change rate of the lateral position deviation as an input feature. Furthermore, a transformer-based surrounding HDVs trajectory-prediction model is developed based on the predicted merging behavior. In the transformer model, we use a sliding window approach to continuously predict the HDV merging trajectories. The prediction accuracy was evaluated across various time windows, and optimal time windows were determined for normal and aggressive driving. For normal driving, the optimal time window was found to be 6.67 seconds, with an MSE of 0.0314. For aggressive driving, the optimal time window was determined to be 11.67 seconds, with an MSE of 0.028. The prediction horizon of the transformer model (with MSE smaller than 0.05) was 3.5 times longer than that of the prevailing LSTM model. The merging trajectory can be precisely predicted up to 30 s in advance, demonstrating the great potential of the transformer prediction model in terms of prediction accuracy and horizon.

Finally, an enhanced ADS motion-planning approach is developed by integrating the predicted HDVs driving style and trajectory. The proposed approach was tested for different prediction horizons. The safety performance was evaluated, achieving TTC values of 300.4 and 383.88 values with the optimal prediction horizon for normal and aggressive driving styles, respectively. Compared with ADS path planning without trajectory prediction of surrounding HDVs, the TTC values of the proposed approach were higher, proving that our proposed approach enables the ADS to reduce the time to collision with surrounding HDVs. In the future, the employment of the proposed approach will be explored in three or four lanes for additional surrounding vehicles. The transferability of the proposed method will be tested for a much more complex urban scenario. The comfort of ADS motion planning will also be investigated.